\documentclass[letterpaper]{article} 
\usepackage{aaai25}  
\usepackage{times}  
\usepackage{helvet}  
\usepackage{courier}  
\usepackage[hyphens]{url}  
\usepackage{amsmath}
\usepackage{amsfonts}
\usepackage{graphicx} 
\usepackage{booktabs}
\usepackage{color}
\usepackage{xspace}
\usepackage{natbib}  
\usepackage{caption} 
\usepackage{algorithm}
\usepackage{algorithmic}
\usepackage{enumitem}
\usepackage{amsthm}
\usepackage{amsmath}
\usepackage{dsfont}

\usepackage{newfloat}
\usepackage{listings}
\newtheorem{myPro}{\textbf{Proposition}}
\newcommand{\method}{\texttt{SODor}\xspace }
\newcommand{\argmin}[1]{\underset{#1}{\mathop{\rm argmin ~}}}

\newcommand{\loglike}{ll}

\newcommand{\tensor}{\mathcal{X}}

\newcommand{\element}{x}

\newcommand{\probdim}{D}

\newcommand{\ndim}{N}
\newcommand{\timestep}{T}

\newcommand{\sparseparam}{\lambda} 

\newcommand{\matXN}[2]{mat(#1)^{ (#2) }}

\newcommand{\TTS}{TTS\xspace}

\newcommand{\Norder}[1]{$(#1)^{th}$-order\xspace}

\newcommand{\lone}{$\ell_1$\xspace}
\newcommand{\lonenorm}{\lone-norm\xspace}

\newcommand{\eq}[1]{Eq.~(#1)}

\newcommand{\model}{\theta} 

\DeclareCaptionStyle{ruled}{labelfont=normalfont,labelsep=colon,strut=off} 
\floatstyle{ruled}
\newfloat{listing}{tb}{lst}{}
\floatname{listing}{Listing}
\title{
Long-Term EEG Partitioning for Seizure Onset Detection}
\author{
   Zheng Chen\textsuperscript{\rm 1},
    Yasuko Matsubara\textsuperscript{\rm 1}, 
    Yasushi Sakurai\textsuperscript{\rm 1}, 
    Jimeng Sun\textsuperscript{\rm 2,3}
}
\affiliations{
    \textsuperscript{\rm 1}SANKEN, Osaka University\\

    \textsuperscript{\rm 2}University of Illinois Urbana-Champaign\\
    \textsuperscript{\rm 3}Carle Illinois College of Medicine, University of Illinois Urbana-Champaign\\
    \{chenz, yasuko, yasushi\}@sanken.osaka-u.ac.jp;
    \indent 
    jimeng@illinois.edu
}

\usepackage{bibentry}

\newtheorem{definition}{Definition}
\newtheorem*{problem}{Problem}

\newcommand{\vx}{\boldsymbol{x}}
\newcommand{\vy}{\boldsymbol{y}}
\newcommand{\vz}{\boldsymbol{z}}
\newcommand{\va}{\boldsymbol{a}}

\begin{document}

\maketitle

\begin{abstract}

Deep learning models have recently shown great success in classifying epileptic patients using EEG recordings. Unfortunately, classification-based methods lack a sound mechanism to detect the onset of seizure events. In this work, we propose a two-stage framework, \method, that explicitly models seizure onset through a novel task formulation of subsequence clustering. Given an EEG sequence, the framework first learns a set of second-level embeddings with label supervision. It then employs model-based clustering to explicitly capture long-term temporal dependencies in EEG sequences and identify meaningful subsequences. Epochs within a subsequence share a common cluster assignment (normal or seizure), with cluster or state transitions representing successful onset detections. Extensive experiments on three datasets demonstrate that our method can correct misclassifications, achieving 5\%-11\% classification improvements over other baselines and accurately detecting seizure onsets.

\end{abstract}

%

\section{Introduction}

\begin{figure}[t]
\centering
\includegraphics[width=0.99\columnwidth]{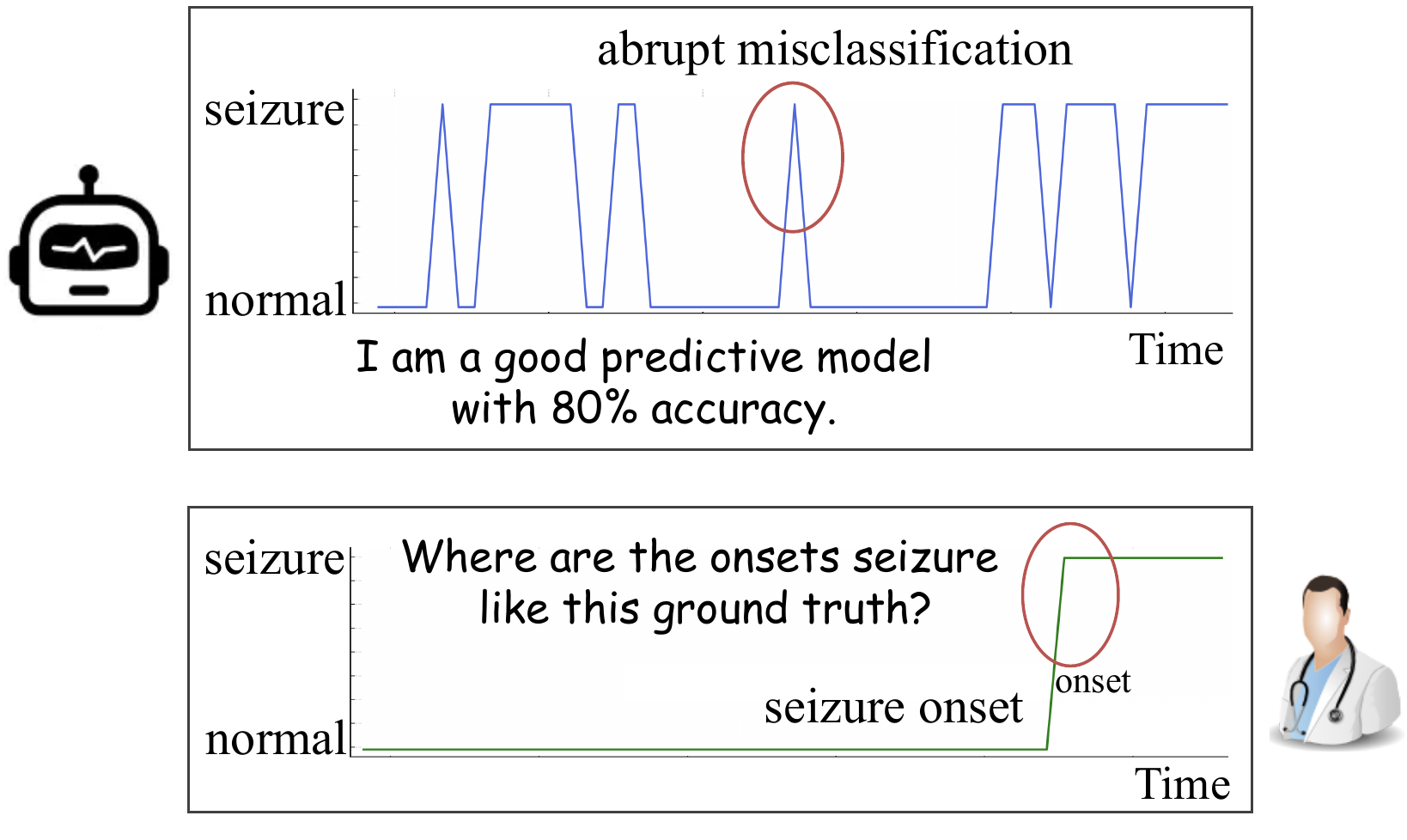}
\caption{A visualization of 55 two-second epochs using an second-level classification method shows some unexpected abrupt misclassifications. 
This issue may lead to unexplainable outcomes for clinicians.}
\label{fig:story}
\end{figure}

Epilepsy affects 60 million of the population worldwide, and approximately 40\% of patients have drug-resistant epilepsy with recurrent seizures that cannot be controlled by available medications \cite{WHO}. 
This problem leads to an increased risk of sudden death.
Deep learning models have demonstrated impressive success in automating seizure detection using electroencephalogram (EEG) data \cite{yang2023biot, GNN_AAAI23, Chen_SDM23}.
Successful methods typically form a classification task.
They divide EEG recordings into a sequence of second-level epochs and aim to classify them accurately.
While state-of-the-art (SOTA) performance has been demonstrated, \emph{these methods cannot provide inherent information of seizure detection research, i.e., the seizure onset (SO).}
Clinically, many epileptic patients benefit from accurate SO detection as it helps localize and surgically remove the onset zone in the brain, which exhibits the earliest electrophysiological changes during a seizure event. 
Also,  successful detection provides optimal timing to adjust abnormal electrical activities in the brain by neuromodulatory devices \cite{OxfordBrain2019}.
Despite great importance, the objective of classification-based methods is to determine whether a seizure exists within an EEG, without explicitly modeling and showing SO position over a long sequence.
Several abrupt misclassifications randomly appear within a state-consistent sequence, as shown in Figure \ref{fig:story}.
These misclassifications inevitably increase false alarms of SO detection and lead to unexplainable outcomes.

A few studies propose a two-stage method \cite{PostProcess2, PostProcess3, PostProcess5, PostProcess4}.
They employ post-processing to smooth abrupt changes and reduce the false detection rate.
Samples within a sliding window are re-assigned a consistent label based on a majority vote. 
However, several limitations remain unsolved.

\begin{itemize}[left=0pt]    
    \item \textbf{Still lacking explicit SO detection modeling.}
    The objective of post-processing is to smooth misclassifications.
    No existing works effectively formulate SO detection as a learning task and directly output the SO information.
    They are required to manually set parameters, e.g., voting threshold and window size, which hinder scalability.

    \item \textbf{Insufficient feature utilization.}
    Existing works estimate a simple statistical observation of label assignments. They ignore the underlying features that characterize an EEG sample and its relation to a seizure event. 
    Such empirical observations fail to offer any explanation for detection.

    \item \textbf{Lacking long-sequence dependencies modeling.}
    Either classification-based methods or voting operations treat all second-level epochs equally and uniformly.
    They fail to take into context information or long-term dependencies within a sequence. This often yields suboptimal results.

\end{itemize}

To tackle these, we propose a new \underline{\textbf{s}}eizure \underline{\textbf{o}}nset \underline{\textbf{d}}etect\underline{\textbf{or}}, \method, with the following contributions:

\begin{itemize}[left=0pt]
    \item \textbf{Subsequence clustering formulation for explicit SO detection.}
    We propose a deep clustering method to explicitly model the SO detection task, which is formulated as a subsequence clustering problem to \emph{find} and \emph{segment} the subsequences that characterize a consistent state (normal vs. seizure) automatically.
    The segment points show state transitions that are identified as SO timestamps.

    \item \textbf{Channel logits representations.}
    We propose a channel correlation representation and successfully formulate it as the clustering objective. 
    This approach benefits from the classification model by leveraging logits of label assignments from EEG channels, rather than smoothing label assignments.
    By learning the correlations between these logits, the method provides insights into \emph{how multi-channel interactions relate to seizures.}

    \item \textbf{Temporal consistency modeling.}
    We propose modeling time-invariant interactions within epochs and long-sequence consistency through a clustering constraint.
    This approach helps mitigate abrupt changes and encourages neighboring epochs to be assigned to the same cluster.
    
\end{itemize}

\noindent To our knowledge, we are the first to formulate SO detection as a clustering task and explicitly output SO information.

\section{Related Works}\label{sec:relatedworks}
\subsection{Seizure Onset Detection}
\label{subsec:SOD}
Existing SO detection methods can be categorized into end-to-end and two-stage approaches.
End-to-end methods frame SO detection as a classification task, labeling second-level epochs as either normal or seizure.  
Various deep learning models such as CNNs \cite{TSTCC_ijcai21,onsetclustering2}, Transformers \cite{BrainNet_KDD22,chenIJCAI,yang2023biot,kotoge2024splitsee}, and Graph models \cite{GNN_ICLR22,GNN_AAAI23,MBrain_KDD23} are employed to automate feature extraction and classification. 
While an accurate model may provide some SO information, these methods do not explicitly model and detect SOs.
Some studies set up onset EEG fragments as a separate class for three-way classification \cite{RasheedIEEETNSRE2021,Dissanayake,Chen_SDM23}. 
Since they focus only on classification accuracy, end-to-end methods often result in abrupt misclassifications at random timestamps, increasing false SO detections.
Two-stage methods involve post-processing the outputs of classification models. 
They re-assign consistent state labels to sequences of epochs within a window based on majority voting \cite{PostProcess2,PostProcess3,PostProcess4} or thresholding \cite{PostProcess1,PostProcess5,onsetclustering2}.
However, these works lack explicit learning for SO detection and oversimplify label smoothing, resulting in suboptimal performance that often requires data-specific manual adjustments.
Different from existing methods, we propose a novel formulation of SO detection as a time series subsequence clustering task.

\subsection{Time Series Subsequence Clustering}
\label{subsec:TS subsequenceclustering}
Subsequence clustering is an important task in time series data mining.
The objective is to distinct states in sequences of time series, without relying on known labels and segments \cite{AutoPlait,Koheiwww2024}.
For example, in a dance routine, a multivariate time series captures transitions between motion states such as “walk,” “run,” “jump,” and “kick.” 
Subsequence clustering segments these time observations into concise segments and assigns each segment a motion label.
Model-based methods are commonly used, where each cluster is represented as a model, and sequences are fitted to these models \cite{TICC,KokiNeurIPS2021,cubescope,Kokiwww}.
Recent researchers propose some deep clustering involving two stages \cite{HVGH2019,Time2State,E2Usd}.
The first stage learns a latent embedding by deep learning models.
Then, they view a sequence of embeddings with multiple dimensions as a multivariate time series, and a clustering stage identifies the state of embeddings.
\method follows this line but differs from these methods, which use unexplainable embeddings for clustering.
Inspired by \cite{OxfordBrain2019,NatNeuroscience2021}, we propose an explainable method that treats the learned channel-independent logits as multivariate time series and models their state-specific structures.

\begin{figure*}[t]
\centering
\includegraphics[width=0.95\linewidth]{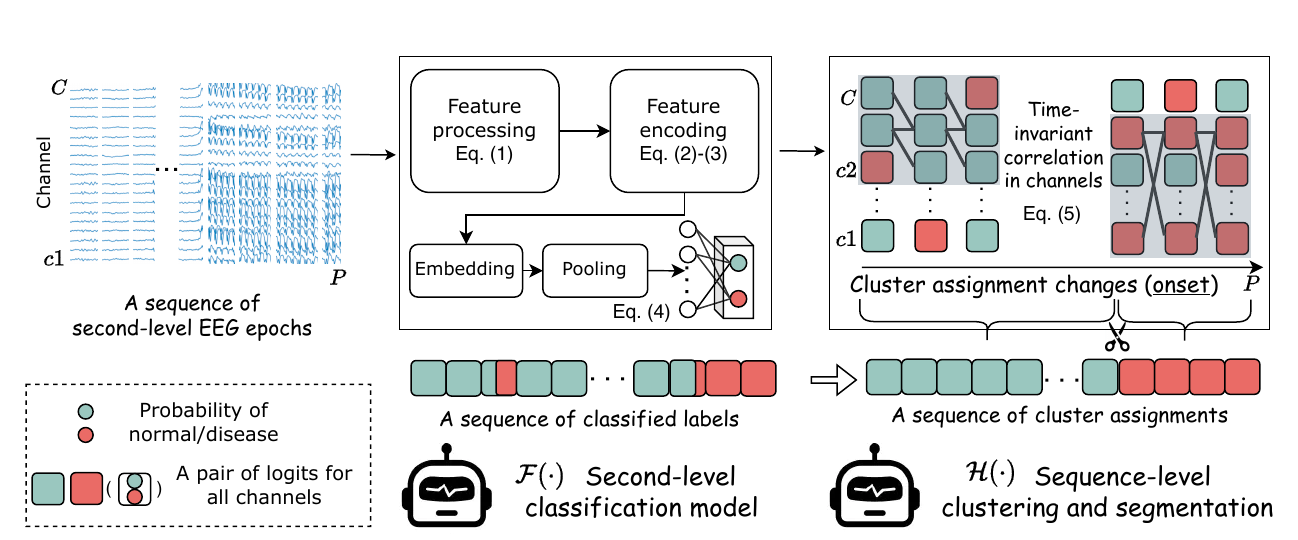}
\caption{
System overview: \method is a two-stage framework designed to explicitly detect seizure onset, consisting of a classification model ($\mathcal{F}(\cdot)$) and a subsequence clustering model ($\mathcal{H}(\cdot)$). $\mathcal{F}(\cdot)$ learns second-level correlation representations for channel-wise logits in normal and seizure states through supervised learning. $\mathcal{H}(\cdot)$ then clusters a sequence of second-level epochs into subsequences. Epochs in a subsequence are consecutive, and each subsequence is dependent on its neighbors. 
A transition between subsequence/cluster assignments can be viewed as a seizure onset. 
}
\label{fig:method}
\end{figure*}

\section{Problem Formulation}\label{sec:problemformulation}

This section presents the problem formulation for SO detection using EEG data. We first introduce preliminaries.

\begin{definition}[\textbf{Multi-channel EEG recordings}]

EEG data captures neuronal activities from different brain regions over time.
Let $\mathcal{X} := {\{\mathcal{X}^{(n)}\}}_{n=1}^{N}$ denote a longitudinal EEG set of $N$ patients.
For $n$-th patient, $\mathcal{X}^{(n)} \in \mathbb{R}^{C \times T}$ represents several recordings from $C$ channels over a duration of $T$ time points.
\end{definition}

To prepare for detecting the SO timestamps, each patient recording is segmented into a sequence of second-level epochs by a sliding window, denoted as $\mathbf{X} = [\mathbf{X}_{1},\mathbf{X}_{2},\dots,\mathbf{X}_{P}]$, each $\mathbf{X}_{p} \in \mathbb{R}^{C \times L}$ and $L$ is the window size.
Each $\mathbf{X}_{p}$ is associated with a label $\vy_p\in\{0,1\}$.

\begin{definition}[\textbf{Subsequence clustering}]\label{def:2}

Suppose $\mathbf{X} = [\mathbf{X}_{1}, \mathbf{X}_{2}, \dots, \mathbf{X}_{P}]$ is a set of multivariate time series.
Subsequence clustering wants to group these $P$ time steps or epochs into $K$ clusters by:

\begin{itemize}[left=0pt]
    \item finding a set of $M$ non-overlapping subsequences of $\mathbf{X}$,\\
    i.e., $\mathbf{\tilde{X}}= [\mathbf{\tilde{X}}_{1}, \dots, \mathbf{\tilde{X}}_{M}]$, where $M \ll P$.
    \item estimate the model representations of $K$ clusters,\\
    i.e.,$ \mathbf{\Theta} = \{\Theta_1, \dots, \Theta_K \}$.
    \item assign each subsequence to one cluster, $\Theta_k$\\
    i.e.,$\tilde{\vy}= \{\tilde{y}_1, \dots, \tilde{y}_{K}\}$ where $\tilde{\vy}_k \subset \{1,\ldots,P\}$.
\end{itemize}
\end{definition}

Epochs in a subsequence $\mathbf{\tilde{X}}_{m}$ are consecutive and each subsequence is dependent on its neighbors.
That is, $\mathbf{\tilde{X}}_{m} = \{t_s, t_e\}$ represents the starting and ending timestamps, serving SO detection.
Moreover, model-based methods define each cluster $\mathbf{\Theta}$ by statistical models, such as Markov chains \cite{cubescope} or Gaussian \cite{AutoPlait}.
This method provides interpretability to each cluster's attributes and the reasoning behind the assignments.

\noindent\textbf{Deep subsequence clustering} introduces a feature learning stage before clustering. 
It projects $\mathbf{X}$ into a set of feature vectors $\mathbf{Z} = \{\vz_{1}, \dots, \vz_{P}\}$.
Each $\vz_{p}$, with $d$ dimensions, is treated as an observation from $d$ sensors or channels.

\begin{problem}[\textbf{Seizure onset detection}]
Given a recording of segmented EEG epochs $\mathbf{X}$, seizure detection aims to predict whether a seizure exists.
In contrast, SO detection seeks to identify the interval of a seizure event, including the starting ($s$) and ending ($e$) positions, represented as $\mathcal{O} = \left\{ (\mathbf{X}_{s}, \mathbf{X}_{e}) \ \middle|\ 1 \leq s < e \leq P \right\}$, where $(\mathbf{X}_{s-1}, \mathbf{X}_{s})$ marks the transition from a normal state (0) to a seizure state (1).

\end{problem}

Typically, this problem is solved by a classification model with post-processing.
Our solution, however, is to develop an integrative framework with two distinct learning models. Conceptually, we formulate this problem as a deep subsequence clustering task:
\begin{enumerate}[left=0pt]
    \item $\mathcal{F}(\cdot): \mathbf{X} \rightarrow \{0, 1\}^{(y)}$, which classifies each epoch, resulting in a set of embeddings $\mathbf{Z}$ that encapsulates the second-level label information $\vy= \{y_1, \dots, y_P\}$.
    \item $\mathcal{H}(\cdot): \mathbf{Z} \rightarrow \tilde{\vy}$, which partitions a set of subsequences for a sequence and assigns each subsequence a state label, $\tilde{\vz}= \{\tilde{\vz}_1, \dots, \tilde{\vz}_{M}\}$ and again $M \ll P$.
\end{enumerate}
For a seizure interval (i.e., a subsequence), we aim to achieve $\mathcal{O} = \left\{ \tilde{\vz}_{s}, \tilde{\vz}_{e} \right\}$ where  
$\forall \tilde{y} \in \mathcal{O}$, $\tilde{y} = 1$ and $\tilde{\vz}_{s}$ explicitly provides SO timestamp.
However, existing methods use unexplainable feature embeddings as the cluster targets (as described in Definition \ref{def:2}).
In contrast, we aim for the discovered subsequence clusters to be interpretable because, in epileptic research, it is often more important to learn discriminative and interpretable patterns that reflect seizures.

\section{Proposed Method}
\label{sec:method}

This section presents our \method framework, as shown in Figure \ref{fig:method}. 
Specifically, it comprises two phases: second-level representation learning and sequence-level clustering aimed at explicitly detecting SO in EEG recordings.

\subsection{Second-Level  Representation Learning}
\label{subsec:secondlevelmodel}

Seizure is fundamentally a network disease \cite{PNAS2014}.
The goal of this graph model is to characterize this network and show how it is relevant to a seizure event.
A network constructed from EEG data $\mathbf{X} = [\mathbf{X}_{1},\dots,\mathbf{X}_{P}]$ can be represented as a graph $\mathcal{G} = (\mathcal{V}, \mathcal{E})$ spanning all $P$ epochs. 
$\mathcal{V}$ denotes the set of vertices, corresponding to the EEG channels, and
$\mathcal{E}$ represents the set of edges, which capture correlations between channels. 
An edge is defined as $e_{ij} = (v_i, v_j)$, where $i, j \in C$.

\subsubsection{Channel Correlation for Graph Construction}

To construct a graph, we extract frequency features in each channel as the node embeddings. 
We apply the Fast Fourier Transform (FFT) to each normalized epoch $\mathbf{X}_p$ to decompose the signal into its frequency spectrum. 
The absolute values, $\vx_p = \{\vx_p^{(1)}, \dots, \vx_p^{(|V|)}\}$, are then extracted to represent the node set $\mathcal{V}$.
For edge initialization, we employ a dynamic connectivity modeling method \cite{GNN_ICLR22}.
We compute the absolute value of the normalized cross-correlation between embeddings $(\vx_p^{(i)}, \vx_p^{(j)})$ of each node pair $(v_i,v_j)$. 
Specifically, we calculate each edge weight $e_{ij}$ as follows:
\begin{equation}
e_{ij} := 
\begin{cases} 
\vx_p^{(i)} \circ \vx_p^{(j)} & \text{if } v_j \in \mathcal{V_\text{top}}(v_i) \\
0 & \text{otherwise}
\label{eq:1}
\end{cases}
\end{equation}
where $\circ$ denotes the normalized cross-correlation, and $\mathcal{V}_\text{top}$ represents the set of neighbor nodes with the highest correlation scores. 
This results in a sparse graph representing each second-level EEG epoch for subsequent feature encoding.

\subsubsection{Channel Logits Representation Learning}
While \method follows the deep clustering paradigm, unlike existing works, it generates explainable feature embeddings as the target of subsequent modeling.
Specifically, we propose learning a set of channel-wise logits. 
These logits encode the information from the original data and represent the probabilities of seizures in various channels. 
That is, we shift the observation from a Boolean value (in post-process methods) and unexplainable embeddings to probabilities, offering insights into ``how multi-channel interactions in an EEG epoch relate to a seizure event."
To learn the logit representations, $\mathcal{F}(\cdot)$ is designed to capture the spatial and temporal dependencies within each epoch.
Given $\mathbf{X}_{p}$ with a processed graph structure, it first learns the spatial information by:
\begin{equation}
\mathbf{Z}_{p}^{\text{spatial}} = \text{ReLU}\left( f_{\text{conv}}\left( \sum_{k=0}^{K} \theta_k \cdot (D^{-1}V_p)^k \mathbf{X}_{p} \right) \right),
\label{eq:2}
\end{equation}
where $V_p$ denotes the adjacency matrix defined by Eq. \eqref{eq:1} and $D$ is the diagonal degree matrix.
$f_{\text{conv}}(\cdot)$ denotes a diffusion convolution with a ReLU activation function, referring to \cite{GNN_ICLR22}.
Hence \( (D^{-1}W)^k \) represents the diffusion process across \( k \) steps.
\( \theta_k \) are trainable parameters.
$\mathbf{Z}_{p}^{\text{spatial}} \in \mathbb{R}^{C \times S}$ where $S \ll L$, is a set of channel-wise feature embeddings. 
For each channel, we aggregate frequency messages $\vx_{p}^{(j)}$ across edges by $\frac{1}{|\mathcal{V}_\text{top}(i)|} \sum_{j \in \mathcal{V}_\text{top}(i)} \vx_{p}^{(j)}$
where \( \mathcal{V}_\text{top}(i) \) represents the set of neighbors of the node $i$ defined by normalized cross-correlations.
Afterward, we use a recurrent neural network to model temporal dependencies along with $S$ time steps:
    \begin{equation}
    \{\vz_{p}^{(1)}, \dots, \vz_{p}^{(C)}\}= \sigma(\text{GRU}(\mathbf{Z}_{p}^{\text{spatial}} )),
    \end{equation}

The $c$-th channel $\vz_{p}^{(c)}$ contains a pair of logits for normal or seizure, obtained by using a softmax $\sigma(\cdot)$ to the hidden state of the last time step. 
A max-pooling layer selects the channel with the maximum logit for loss calculation:
\begin{equation}
\mathcal{L}_{\text{BCE}} :=  -\left[ y \log(\vz_{p}^{\text{max}}) + (1 - y) \log(1 - \vz_{p}^{\text{max}}) \right]
\label{eq:bce}
\end{equation}
where $\vz_{p}^{\text{max}}$ represents the maximum logit value across all channels.
Pooling operations are commonly used to aggregate node features, with each method making different assumptions about the graph structure.
We assume that max-pooling retains only the most prominent signals, which may lead to better performance, particularly in tasks where certain strong features are indicative of a seizure event. This assumption aligns with findings suggesting that a few unique, abnormal connections across EEG channels can serve as SO markers \cite{NatNeuroscience2021,PostProcess5}.
A comprehensive evaluation is provided in the Experiment section.

\subsection{
Sequence-Level Clustering for SO Detection}\label{subsec:SOdectection}

\subsubsection{Formulating SO Detection as A Subsequence Clustering}
Given second-level representations $\mathbf{Z} = \{\vz_{1}, \dots, \vz_{P}\}$ with $C$ multivariate sequences, our goal is to cluster and segment them into subsequences $\tilde{\mathbf{Z}}= \{\tilde{\mathbf{Z}_1}, \dots, \tilde{\mathbf{Z}}_{M}\}$.
As described in Definition \ref{def:2}, each subsequence contains several EEG epochs with consistent clustering, and a pair $\{\tilde{\mathbf{Z}}_m,\tilde{\mathbf{Z}}_{m+1}\}$ represents a cluster transition, which facilitates SO detection.
In this work, $\mathcal{H}(\cdot)$ is designed on top of a novel Toeplitz Inverse Covariance-based Clustering \cite{TICC}.
This method employs a graphical lasso to estimate sparse Gaussian inverse covariance matrices, also known as precision matrices \cite{Koheiwww2024}, to represent cluster models $\{\Theta_k | k = \text{normal}, \text{seizure}\}$. 
Each matrix provides insights into pairwise conditional independencies among EEG channels, determining which correlations contribute most significantly to cluster assignments.
The formulation is:
\begin{align}
\argmin{\mathbf{\Theta}, \tilde{\vy}} &\sum_{k=1}^{K} \Bigg[
 \sum_{y =1 }^{|Y|} \sum_{\vz_p \in \tilde{\vy}_k}  (\underbrace{-\ell\ell(\vz_p,\Theta_k)}_{\text{\normalfont log likelihood}} \\ \nonumber
 &+ \underbrace{\beta \mathds{1}\{\vz_{p-1} \not\in \tilde{\vy}_k\}}_{\text{\normalfont temporal consistency}} ) + \underbrace{\|  \lambda \odot \Theta_k\|_1}_{\text{\normalfont time-invariant representation}}\Bigg].
\label{TICC}
\end{align}
where the log-likelihood term measures the probability that the $p$-th epoch belongs to cluster $k$ by observing its representation $\vz_{p}$.
The temporal consistency term models long-term dependencies, encouraging neighboring epochs to be assigned to the same cluster.
$\|\lambda \odot \Theta_k\|_1$ is an $\ell_1$-norm penalty to control the sparseness of $\Theta_k$.
This enforces the preservation of cluster-specific correlations between EEG channels, enabling the learning of time-invariant representations.

\noindent \textbf{- \emph{Limitation.}} Notably, when $|Y| = 1$, Eq. (5) denotes the original clustering algorithm \cite{TICC}.
However, while the original clustering algorithm operates on multivariate time series as input, our pairwise logit representation is structured as a two-dimensional tensor time series.
Next, we address how to define and infer the model  $\Theta_k$.

\subsubsection{Logits Toeplitz Matrices}
Instead of clustering each epoch independently, we assume neighboring epochs should be consecutive, so we redefine the ``epoch" by a sliding window $\omega \ll P$, represented as $\mathbf{Z}_p := \{\mathbf{Z}_{p-\omega+1}, \dots, \mathbf{Z}_p\}$. 
Thus, we cluster these short-duration matrices and then fit all variables into $\Theta$, characterized by block Toeplitz inverse covariance matrices.
These block-wise constraints are designed to capture time-invariant structural patterns within $\mathbf{Z}_p$, helping to smooth abrupt changes. A matrix can be expressed as:

\[
\small
\Theta_k := 
\left[
\begin{array}{c c c c c c}
A^{(0)} & (A^{(1)})^P  & \cdots & \cdots & (A^{(w-1)})^P \\
A^{(1)} & A^{(0)} &  \ddots & & \vdots \\
A^{(2)} & A^{(1)}  & \ddots & \ddots & \vdots \\
\vdots & \ddots  & \ddots & (A^{(1)})^P & (A^{(2)})^P \\
\vdots &  & A^{(1)} & A^{(0)} & (A^{(1)})^P \\
A^{(w-1)} & \cdots  & A^{(2)} & A^{(1)} & A^{(0)} \\
\end{array}
\right],
\label{eq:matrix}
\]
where each $A$ represents logit correlations among $C$ channels within $\omega$ time observations. 
An element $a_{i,j} \in A$ refers to the relationship between the $i$-th and $j$-th channels at the same $\omega$-th epoch or between the $(\omega-1)$-th and $\omega$-th epochs.

However, the original graphical lasso estimates $\Theta_k$ based on single observations $T$.
Instead, we formulate a pair of logits as a two-dimensional tensor, treating it as a single observation. 
Such estimations can become too high-dimensional. 
\cite{Kokiwww} proposes separating tensor $\Theta_k$ into multimode, where $\va^{(n)}_{i,j} \in A^{(n)}$ refers to the relationship between the $i$-th and $j$-th variables in mode-$n$. 
This may lead to over-representation, since the logit for "normal" already implies the probability of "seizure".
Since $\Theta$ are covariance matrics, we solve this by the linearity property of covariance \cite{linearliy}.

\begin{myPro}
\label{Theo: norm in time doamin}
Given a pair of logits $\{\vz_\text{nor},\vz_\text{sei}\}$, denoted as $\{A,\tilde{A}\}$, under the constraint $\vz_\text{nor} + \vz_\text{sei} = 1$, computing     $\text{Cov}(A, \tilde{A})$, as it fully captures the covariance relationship between $A$ and $\tilde{A}$, due to the linearity property and \(\text{Cov}(A, 1) = 0\), as denoted by $\text{Cov}(A, A) \equiv \text{Cov}\{A,\tilde{A}\}$.

\end{myPro}

The proof is provided in the appendix.
Thus, we focus on analyzing the logit representation of the seizure state, which captures pairwise logit correlations.
This proof allows us to estimate $\mathbf{\Theta}$.
Furthermore, this single logit representation can seamlessly serve as the optimization objective for the BCE loss in Eq. \eqref{eq:bce}.

\subsubsection{Clustering for SO detection}\label{subsubsec:SOdetection}
After estimating the cluster representation $\mathbf{\Theta}$, we identify the optimal $\Theta_k$ for each $\mathbf{Z}_p$. 
More importantly, we incorporate temporal consistency by ensuring that consecutive EEG epochs are aligned to a consistent representation, thereby further modeling long-term dependencies in a state-consistent sequence, as denoted by:

\begin{align}
\small
\mathrm{minimize} \sum_{k=1}^K \sum_{\vz_p \in \tilde{\vy}_k} -\ell\ell(\mathbf{Z}_p,\Theta_k) + \beta \mathds{1}\{{\mathbf{Z}_{p-1} \not\in \tilde{\vy}_k\}}. 
\label{eq:clustAssign}
\end{align}

This formulation jointly maximizes the log-likelihood and maintains temporal consistency. The balance between these objectives is controlled by $\beta\mathds{1}$. 
This is an indicator function: when the same cluster $\tilde{\vy}_k$ is made between neighboring epochs, there is no penalty, but $\mathds{1}\{t - 1 \notin P_k\}$is 1, if $\mathbf{Z}_{p-1}$ does not belong to the same cluster as $\mathbf{Z}_{p}$.

Minimizing this constraint ensures that neighboring EEG epochs are assigned to the same clusters.
Any assignment that deviates from this constraint indicates a state transition.
The corresponding 
$p$-index marks the onset of the next state, identifying SO when transitioning from a normal state to a seizure state in the EEG sequence.

\subsubsection{Optimization}\label{subsubsec:opeimization}
The subsequence clustering is optimized by the expectation-maximization (EM) algorithm to iteratively learn the cluster assignments $\tilde{\vy}$ and the structural patterns $\mathbf{\Theta}$ until convergence. 
Specifically, in Eq. (5), the log-likelihood term and the sparsity term, which can be considered as a typical graphical lasso problem, have a solution guaranteed to converge to the global optimum using the alternating direction method of multipliers (ADMM)~\cite{admm}. 
The clustering is formed by a dynamic programming optimization that finds the minimum cost Viterbi path for a sequence \cite{TICC}.
More detailed methods, implementations, and optimizations can be found in the appendix.

\begin{table*}[t]
    \centering
    \resizebox{0.78\linewidth}{!}{%
    \begin{tabular}{l|lll|ccc|ccc}
    \toprule
         & \multicolumn{3}{c|}{CHB-MIT} & \multicolumn{3}{c|}{HUH} & \multicolumn{3}{c}{TUH} \\ \midrule
         Baseline    & NMI & ARI & ACC & NMI & ARI & ACC & NMI & ARI & ACC \\ \midrule
    Burrello, TBME, 2020 (PP)       & 0.897       & 0.875       & 0.882       & 0.574 & 0.728 & 0.661 & 0.683 & 0.656 & 0.717 \\
       Boony, TSNRE, 2021  (PP)            & 0.842       & 0.744       & 0.868       & 0.732 & 0.715 & 0.749 & 0.670 & 0.692 & 0.713 \\
    Li et al., TSNRE, 2021     (PP)        & 0.840       & 0.859       & 0.870       & 0.643 & 0.625 & 0.639 & 0.617 & 0.604 & 0.699 \\
    Siyi et al., ICLR, 2022    (Cls)      & 0.865       & {0.867}       & {0.939}       & 0.778 & 0.744 & 0.810 & 0.764 & 0.743 & 0.816 \\
    Thi et al., AAAI, 2023   (Cls)        & 0.815       & 0.817       & 0.889       & 0.728 & 0.694 & 0.762 & 0.716 & 0.695 & 0.768 \\
     Time2State, WWW, 2023  (Clu)         & 0.770       & 0.730       & 0.765       & 0.632 & 0.654 & 0.733 & 0.664 & 0.673 & 0.756 \\
     E2Usd, WWW, 2024  (Clu)        & 0.834       & 0.815       & 0.852       & 0.669 & 0.686 & 0.704 & 0.668 & 0.677 & 0.760 \\
    \textbf{\method}             & \textbf{0.979}       & \textbf{0.964}       & \textbf{0.981}       & \textbf{0.873} & \textbf{0.823} & \textbf{0.845} & \textbf{0.842} & \textbf{0.863} & \textbf{0.879} \\
    \bottomrule
    \end{tabular}
    }
    \caption{Performance comparison on the CHB-MIT, HUH, and TUH datasets.
We retain the initial classification model stage and utilize the post-processing and clustering methods from the baseline approaches.
\textbf{Bold}: best; PP: post-process; Cls: one-step classification-based method; Clu: deep clustering method.
    }
        \label{tab:seizure_main}
\end{table*}

\section{Experiments and Results}\label{sec:experiment}
We evaluate \method to determine if it addresses the following questions:

\begin{itemize}[left=0pt]  
    \item Can \method filter out the false detections, e.g., abrupt misclassifications, and can automatically detect the SO.
    \item Are the channel logit correlation representations robust and beneficial for SO detection?
    \item Does \method potentially provide explainable clusters?
\end{itemize}

\subsection{Datasets}\label{subsec:data}

We evaluated \method for the seizure onset (SO) detection task on three real-world datasets.
In addition, we assessed a more challenging task: seizure prediction, which aims to identify the preictal state preceding seizures.
This task is critical in clinical settings, and we provided onset information for it, referred to as PSO (Preictal Seizure Onset).

\noindent\textbf{CHB-MIT} comprises 844 hours of continuous scalp EEG data from 22 patients, recorded across 22 channels, with a total of 163 seizure episodes.
For the PSO detection task, we preprocessed this dataset by defining the pre-seizure state as the 5 minutes preceding seizure onset.

\noindent\textbf{HUH} is collected from University of Helsinki, Finland. It consists of scalp 21-channel EEG data of 79 patients, serving the seizure detection task.

\noindent\textbf{TUSZ} dataset is part of the Temple University Hospital EEG Seizure Corpus. It comprises 5,612 EEG recordings with 3,050 clinically annotated seizures.
We utilized 19 EEG channels, following the standard 10-20 system.

We divided each dataset into 70\%/20\%/10\% for training, testing, and validation. 
We stored the IDs of all epochs in the patient recordings, enabling recall in long-term recordings to verify detection accuracy.

\subsection{Baselines}\label{subsec:baseline}
We compared \method with three post-process (PP) SO detection baselines, two classification-based (Cls) seizure detection baselines, and two deep clustering (Clu) methods.

\begin{enumerate}
    \item \cite{PostProcess2} proposed a post-processing that uses a sliding 5-second window to re-assign labels based on a patient-specific voting threshold.
    \item \cite{PostProcess5} proposed a two-step method involving a deep model to classify second-level epochs first and a weighting phase to score the probabilities of SO.
    \item \cite{PostProcess3} proposed an ensemble learning post-process based on four machine learning models and used majority voting to determine the SO threshold.
    \item \cite{GNN_ICLR22} proposed a GNN-based method for seizure detection and classification tasks.
    \item \cite{GNN_AAAI23} proposed a GNN model incorporating contrastive learning for seizure classification.
    \item Time2State \cite{Time2State} is a deep clustering method tailored for multivariate time series.
    \item E2Usd \cite{E2Usd} formulates each dimension of feature embeddings as a multivariate time series and conducts a subsequence clustering on them.
\end{enumerate}

\subsection{Experiment: Overall SO Detection}\label{subsec:exp1}
\noindent\textbf{Setup.} 
We compared selected baselines to \method. 
For a fair comparison, with \cite{PostProcess2}, \cite{PostProcess5}, and \cite{PostProcess3}, we maintained our first-stage second-level learning and compared their post-processing methods. For one-step classification methods \cite{GNN_ICLR22} and \cite{GNN_AAAI23}, we used their learning models as the backbone and added our subsequence clustering. 
For Time2State and E2Usd, since they are tailored for time series, we used our classification model and incorporated it into their model-based clustering, the Dirichlet Process Gaussian Mixture Model.\\
\noindent\textbf{Metrics.} We evaluated performance using two clustering metrics: normalized mutual information (NMI) and adjusted Rand index (ARI), along with accuracy (ACC) for assignment analysis.\\
\noindent\textbf{Results (main).} 
Table \ref{tab:seizure_main} presents the main SO detection results across three datasets. \method outperforms all baseline methods. 
Specifically, for CHB-MIT, \method achieved an NMI of 0.979, an ARI of 0.964, and an ACC of 0.981, significantly surpassing the performance of other approaches. The post-processing method \cite{PostProcess2} performed well on this dataset, achieving an NMI of 0.897, an ARI of 0.875, and an ACC of 0.882; however, \method clearly demonstrated superior performance.
One-step classification methods outperformed the two deep subsequence clustering methods (Time2State and E2Usd), as these clustering methods lacked a temporal consistency term. Post-processing and deep clustering methods performed worse on the TUH dataset, likely due to the presence of diverse seizure types and rapid transitions \cite{GNN_ICLR22}. These transitions or preictal phases often contain numerous misclassifications that dominate sequences \cite{preictal2019}, presenting significant challenges for majority voting.

\begin{figure}[t]
\centering
\includegraphics[width=\columnwidth]{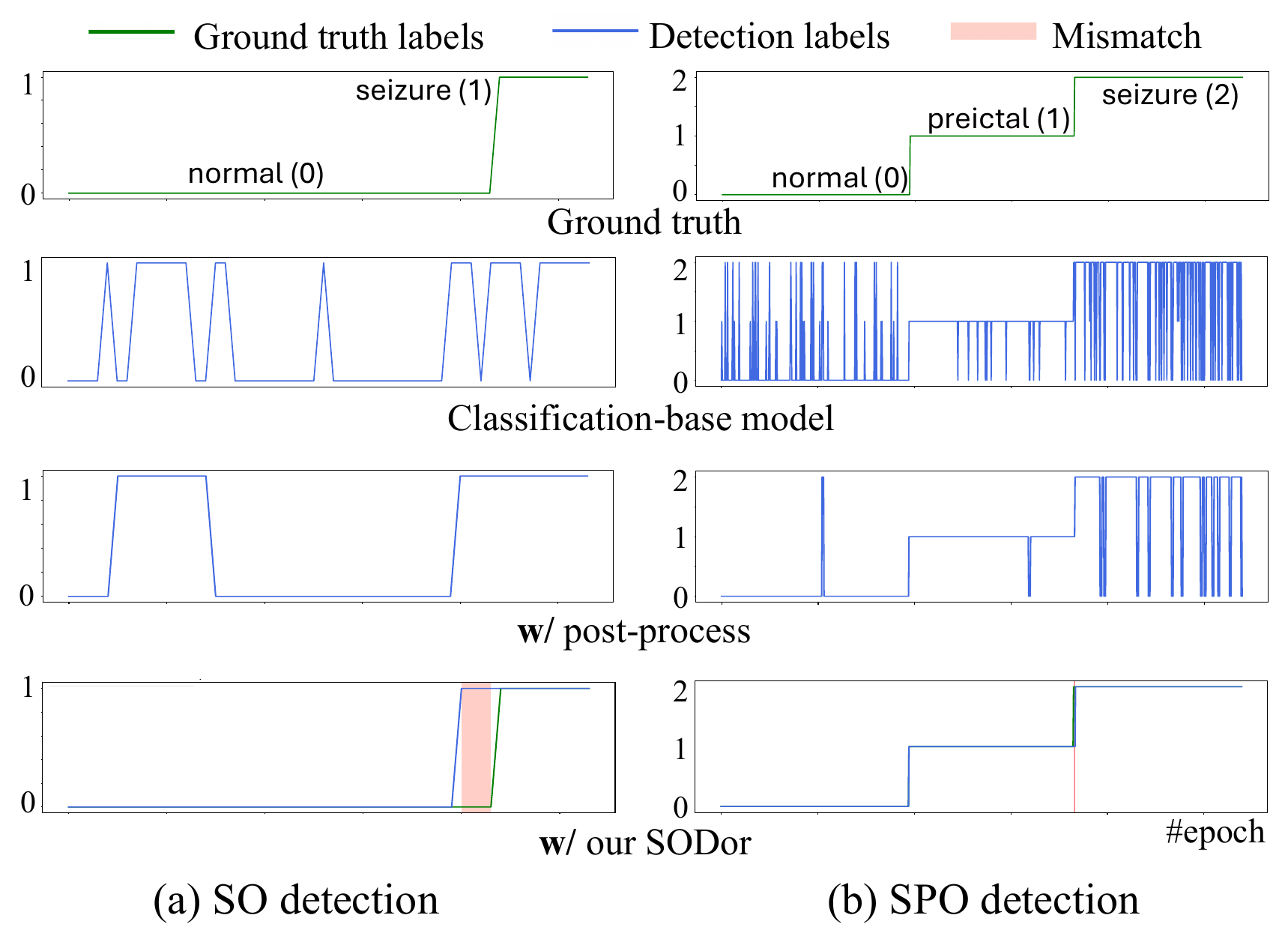}
\caption{Comparison of (a) SO detection and (b) SPO (Seizure Preictal Onset) detection.
The figure illustrates the performance of the classification model (\cite{GNN_ICLR22}), w/ post-processing methods (\cite{PostProcess2}), and w/ our proposed subsequence clustering approach. 
The mismatch between \method and ground truth is marked.}
\label{fig:3}
\end{figure}

\noindent\textbf{Results (visualization).} Figure \ref{fig:3} shows two case study visualizations for the TUH and CHB-MIT datasets. 
Figure \ref{fig:3}(a) shows a sequence with 55 epochs.
The second row shows the results of the SOTA method \cite{GNN_ICLR22}. We applied the post-processing method from \cite{PostProcess2} to the results of the second classification model using a window $\omega = \{2, 3, 5, 7, 10\}$ and a search of voting threshold within $\{0.5, 0.65, 0.75\}$ . 
The final row shows the results with our subsequence clustering.
As shown in Figure \ref{fig:3}(a), although \method exhibits some mismatches between the ground truth and detection labels, it avoids abrupt changes and operates in an automated manner. In contrast, while post-processing reduces false detections, it relies on data-specific manual parameter tuning.
Figure \ref{fig:3}(b) extends the analysis to SPO detection. 
\method also shows significantly fewer mismatches, with minor discrepancies highlighted in the red box.

\begin{figure}[t]
\centering
\includegraphics[width=\columnwidth]{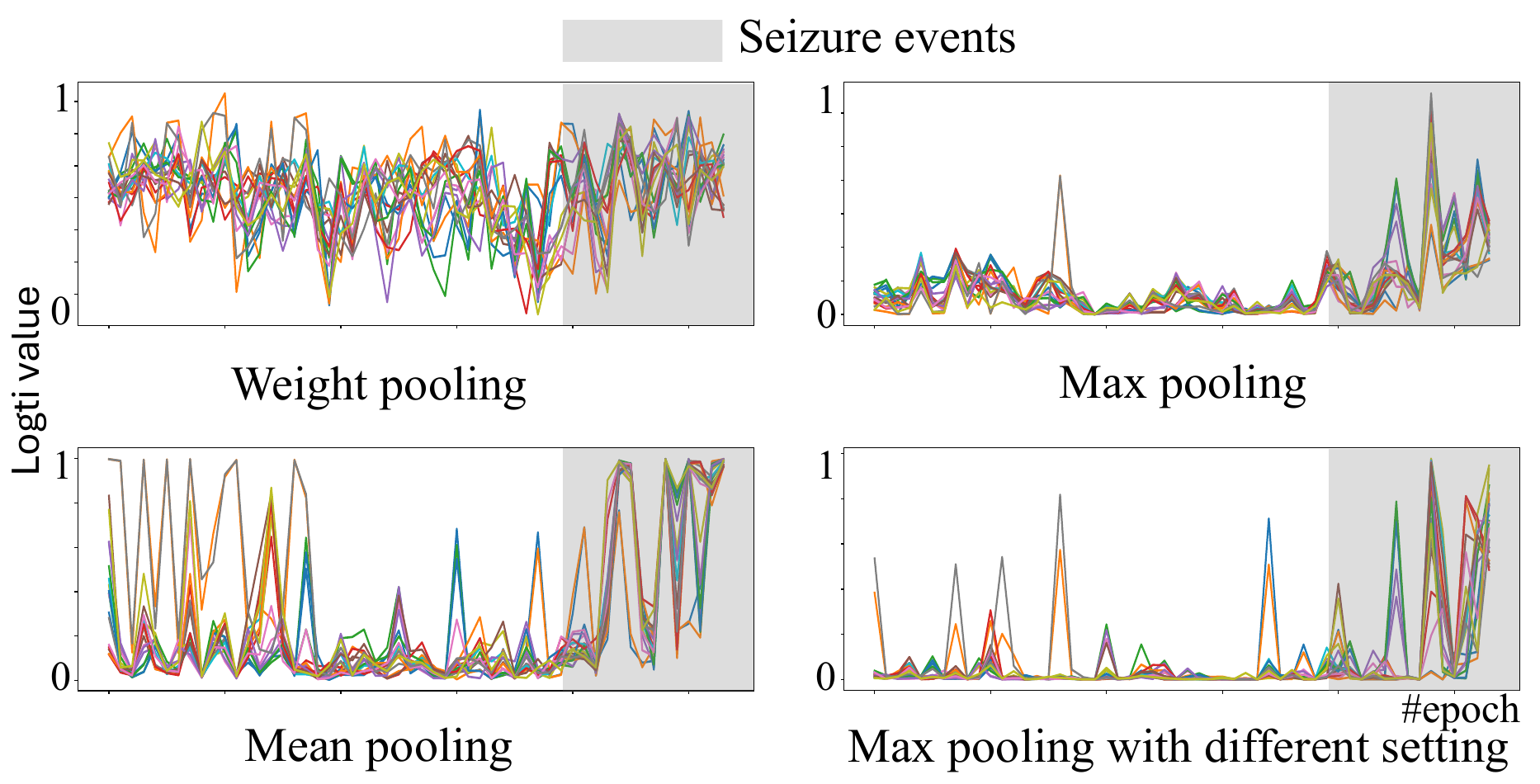}
\caption{Ablation of pooling methods (i.e., max, mean, and weighted) on channel logits across different training settings. 
The results show the robustness of max pooling.
We visualize logits in max pooling under different parameter settings, and it maintains consistent performance.}
\vspace{-5pt}
\label{fig:4}
\end{figure}

\subsection{Experiment: Representation Analysis}\label{subsec:exp2}

\noindent\textbf{Setup.} To maintain consistency in the analysis, we further visualized the TUH case used in Figure \ref{fig:3}(a), focusing on the logits from different channels after training.
As discussed in the ``Proposed Method" section, pooling operations play a critical role in our framework, particularly in summarizing and extracting significant features from channels.
To evaluate their impact, we replaced the proposed max pooling with weighted pooling (using an MLP) and mean pooling, keeping the same parameter settings.
Interestingly, the final classification accuracy remained unchanged at 0.816.\\
\noindent\textbf{Results.} Figure \ref{fig:4} illustrates the impact of different pooling methods on channel-wise logits. It is clear that the representation generated by \method has distinctive characteristics between normal and seizure states. Max pooling exhibits superior robustness compared to other methods. Even though the classification accuracy is the same, weighted pooling distorts the representations, and mean pooling averages the features of graph nodes, which leads to some channels having high probabilities of seizure in normal samples. Moreover, we empirically observe an unstable representation issue during different training settings, but max pooling remains robust across different settings, as shown in the last subfigure.

\subsection{Experiment: Channel Correlation Visualization}\label{subsec:exp3}

\noindent\textbf{Setup.} Since $\mathbf{\Theta}$ represents inverse covariance matrices and the sparsity term in Eq. (5) controls the sparsity of each cluster model, we extracted the sparse matrices as the adjacency matrix after training.
Because our representation preserves the channel index across all learning and clustering stages, we mapped the channels back to the EEG electrodes and graph node positions, following the method outlined in \cite{GNN_ICLR22}, to visualize channel correlations in normal and seizure cluster models.
For CHB-MIT, which contains more than 19 EEG channels, we used the standard 19-channel system for visualization, consistent with the settings for TUH.
Each connection in the visualization corresponds to a ``1" that persists in $\mathbf{\Theta}$.

\begin{figure}[t]
\centering
\includegraphics[width=\columnwidth]{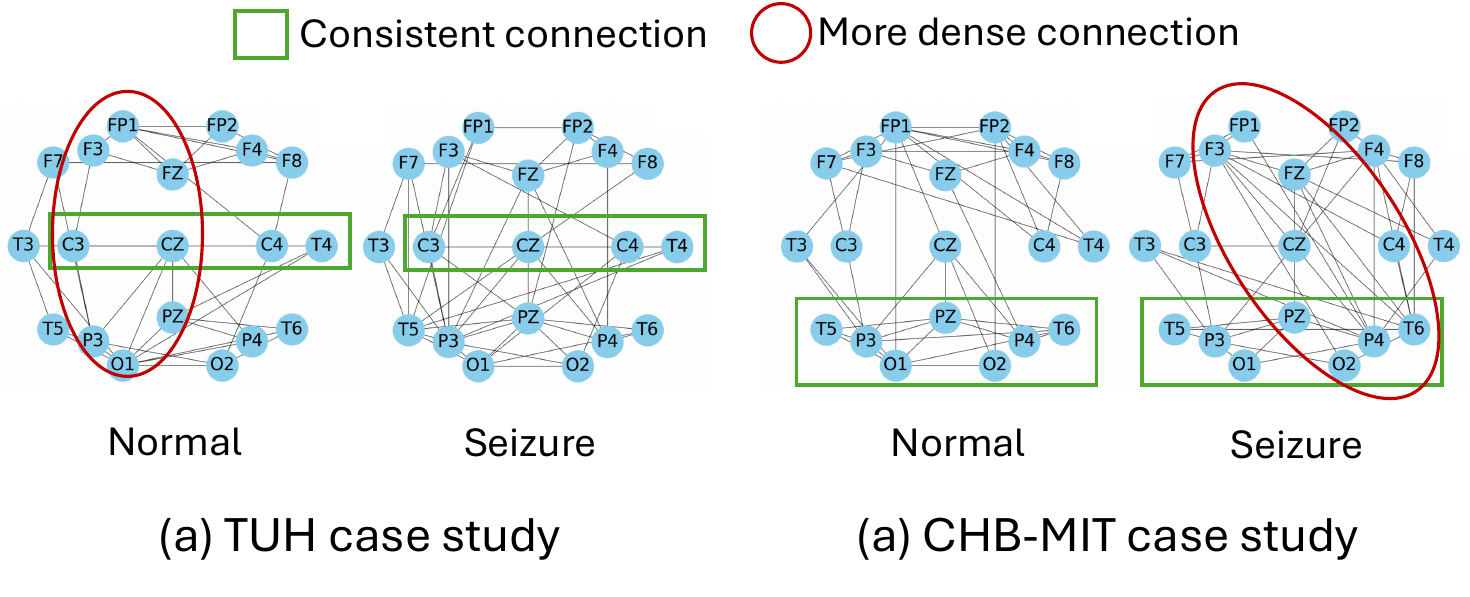}
\caption{Visualization of learned $\mathbf{\Theta}$. 
The connections become denser from normal to seizure.
The green box highlights consistent connections, while the red circle indicates differences between the two clusters.}
\label{fig:5}
\end{figure}

\noindent\textbf{Results.} Figure \ref{fig:5} compares the visualizations of normal and seizure states in two different cases. The results show a consistent pattern: from sparse to dense connections, indicating that more channels are connected and the brain becomes more active during seizures. The central connection of "C3-CZ-C4" in the normal state still appears during seizures, as shown in Figure \ref{fig:5}(a), but more channels are activated. Figure \ref{fig:5}(b) shows that the Occipital lobe (O1 and O2) and Parietal lobe (PZ) remain stable across the transitions, suggesting that the seizure may be related to other areas.

\subsection{Clustering Method Ablation}\label{subsec:exp4}

Tale \ref{tab:clusteringablation} presents that our method can significantly outperform both MDL and DPGMM, achieving the highest values with the lowest standard deviation. 
The reason may be that the MDL-based method \cite{AutoPlait} focuses on model compression loss without explicitly considering temporal consistency. 
DPGMM \cite{E2Usd} focuses more on estimating the number of clusters. 
Our method can also employ Bayesian information criterion (BIC) to estimate the number of clusters. 
Since the seizure states are predetermined, this work specifies it to two clusters for SO detection and three for SPO detection.

\begin{table}[t]
\centering
\begin{tabular}{cccc}
\toprule
    & \method  & MDL & DPGMM \\ \midrule
NMI & \textbf{0.979 $\pm$ 0.04} & 0.720 $\pm$ 0.13    & 0.613$\pm$ 0.09 \\
ARI & \textbf{0.964 $\pm$ 0.04} & 0.694  $\pm$ 0.13   & 0.658$\pm$ 0.09 \\
ACC & \textbf{0.981$\pm$ 0.04} & 0.649  $\pm$ 0.14   & 0.626 $\pm$ 0.09\\ \bottomrule
\end{tabular}
\caption{Ablation study of different subsequence clustering methods: MDL \cite{AutoPlait,Kokiwww} and DPGMM \cite{E2Usd}, in CHB-MIT dataset.}
\label{tab:clusteringablation}
\end{table}



\section{Conclusion}\label{sec:conclusion}

Several powerful seizure detection models have been proposed, yet no existing works explicitly model the seizure onset, often resulting in unexplainable labeling in long-sequence EEG recordings. 
We aimed to provide a robust SO detection framework that successfully formulates this task as subsequence clustering, identifying the state (normal or seizure) transition as the SO timestamp. 
One advantage of this framework is its two-stage learning process, allowing us to fully leverage the high capabilities of existing deep learning methods. Experimental results confirmed that our framework has a strong capacity for SO detection and may have potential for clinical applications.

\section{Acknowledgments}

This work was supported by
JSPS KAKENHI Grant-in-Aid for Scientific Research Number
JP24K20778,    
JST CREST JPMJCR23M3,
and NSF award SCH-2205289, SCH-2014438, and IIS-2034479.
We thank Rikuto Kotoge, Xihao Piao, and Kohei Obata for their valuable discussions.

\bibliography{aaai25.bib}

\newpage
\appendix

\begin{figure}[t]
\centering
\includegraphics[width=\columnwidth]{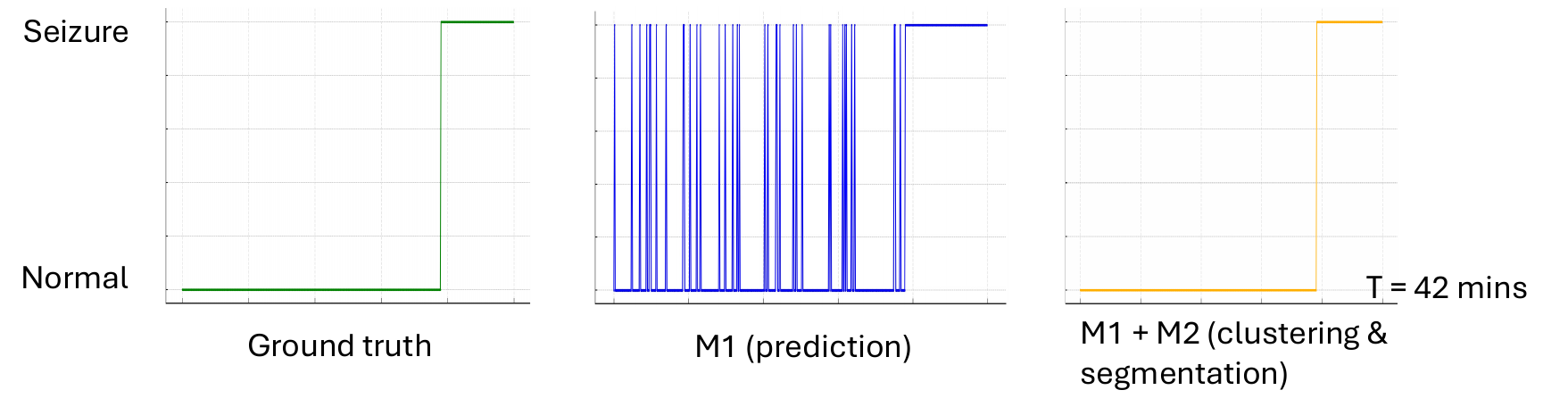}
\caption{Addtional SO detection result}
\label{fig:1}
\end{figure}

\noindent 
Symbols and notations used in this paper can be found in Table \ref{tab:symbols}.
The Appendix is structured as follows:
Appendix \ref{apdx:implement} provides detailed experimental settings, such as hyperparameters, in an anonymous GitHub link.
Appendix \ref{apdx:proof} presents the proof of \emph{Proposition 1} in the manuscript.
Appendix \ref{apdx:graphlasso} describes the optimization process of the subsequence clustering in \method.
Figure \ref{fig:1} shows additional results.

\begin{table}[h!]
\centering
\caption{Summary of Symbols and Notations}
 \resizebox{0.98\columnwidth}{!}{%
\begin{tabular}{|c|l|}
\hline
\textbf{Symbol} & \textbf{Description} \\ \hline
$X^{(n)}$ & Longitudinal EEG set of $N$ patients \\ \hline
$C$ & Number of channels in EEG recordings \\ \hline
$T$ & Duration of time points in EEG recordings \\ \hline
$X_p$ & A second-level epoch from EEG sequence \\ \hline
$L$ & Window size for segmenting EEG recordings \\ \hline
$y_p$ & Label associated with epoch $X_p$ (normal or seizure) \\ \hline
$Z$ & Set of feature vectors derived from EEG data \\ \hline
$z_p$ & Feature vector at epoch $p$ \\ \hline
$M$ & Number of subsequences identified in EEG sequence \\ \hline
$\Theta_k$ & Model representations of clusters \\ \hline
$O$ & Interval of a seizure event with starting and ending positions \\ \hline
$F(\cdot)$ & Classification model for second-level correlation representations \\ \hline
$H(\cdot)$ & Subsequence clustering model \\ \hline
$\lambda$ & Regularization parameter for enforcing sparsity in $\Theta$ \\ \hline
$\beta_1$ & Weight parameter for temporal consistency in clustering \\ \hline
$\text{ReLU}(\cdot)$ & Rectified Linear Unit activation function \\ \hline
$D^{-1}$ & Diagonal degree matrix \\ \hline
$V_p$ & Adjacency matrix defined by channel correlations \\ \hline
$W$ & Matrix representing diffusion process \\ \hline
$\text{GRU}(\cdot)$ & Gated Recurrent Unit used for temporal modeling \\ \hline
$L_{BCE}$ & Binary Cross Entropy loss \\ \hline
$P$ & Number of epochs in the EEG sequence \\ \hline
$y$ & Set of second-level label information \\ \hline
$y_k$ & Cluster assignment \\ \hline
$A^{(0)}$, $A^{(1)}$ & Logit correlations among channels \\ \hline
$\text{Cov}(A, \tilde{A})$ & Covariance relationship between logits \\ \hline
\end{tabular}
}
\label{tab:symbols}
\end{table}

\section{Implementation Details}\label{apdx:implement}

We provide our source code at: \url{https://github.com/chenzRG/SODor}.

\section{Proposition Proof}\label{apdx:proof}  

Given $N$ samples, each containing two real numbers (logits) $A$ and $B$, representing the probabilities of an event, with the constraint $A + B = 1$. We aim to calculate the covariance between samples using the $\arctan$ function.

For each pair of $i$-th and $j$-th sample, we consider the following covariance matrices:

1. $\text{Cov}(A_k, A_j)$
2. $\text{Cov}(A_k, B_j)$
3. $\text{Cov}(B_k, A_j)$
4. $\text{Cov}(B_k, B_j)$

Given $B_k = 1 - A_k$, we have:

\[ \text{Cov}(A_k, B_j) = \text{Cov}(A_k, 1 - A_j) \]

Using the linearity property of covariance:

\[ \text{Cov}(A_k, 1 - A_j) = \text{Cov}(A_k, 1) - \text{Cov}(A_k, A_j) \]

Since \(\text{Cov}(A_k, 1) = 0\), we get:

\[ \text{Cov}(A_k, 1 - A_j) = -\text{Cov}(A_k, A_j) \]

Therefore:

\[ \text{Cov}(A_k, B_j) = -\text{Cov}(A_k, A_j) \]

Similarly, for \(\text{Cov}(B_k, A_j)\):

\[ \text{Cov}(B_k, A_j) = \text{Cov}(1 - A_k, A_j) = -\text{Cov}(A_k, A_j) \]

Finally, for \(\text{Cov}(B_k, B_j)\):

\[ \text{Cov}(B_k, B_j) = \text{Cov}(1 - A_k, 1 - A_j) = \text{Cov}(A_k, A_j) \]

This shows we only need to analyzing $\text{Cov}(A_k, A_j)$ = $\text{Cov}\{A, B\}$

\begin{proof}
\textbf{The covariance between a random variable and a constant is zero.}
Consider a random variable $X$ and a constant $c$. The covariance is given by:

\[
\text{Cov}(X, c) = \mathbb{E}[Xc] - \mathbb{E}[X]\mathbb{E}[c]
\]

Since $c$ is a constant, we have:

\[
\mathbb{E}[Xc] = c\mathbb{E}[X]
\]

And, the expectation of a constant is the constant itself:

\[
\mathbb{E}[c] = c
\]

Thus, the covariance can be written as:

\[
\text{Cov}(X, c) = c\mathbb{E}[X] - \mathbb{E}[X]c = 0
\]

\end{proof}

\begin{proof}
\textbf{The linearity property of covariance} \cite{linearliy}
For two random variables \(A_k\) and \(A_j\), and a constant \(1\), the covariance is given by:

\[ \text{Cov}(A_k, 1 - A_j) = \mathbb{E}[A_k (1 - A_j)] - \mathbb{E}[A_k]\mathbb{E}[1 - A_j] \]

We can distribute and separate the terms:

\[ \mathbb{E}[A_k (1 - A_j)] = \mathbb{E}[A_k \cdot 1 - A_k \cdot A_j] = \mathbb{E}[A_k] - \mathbb{E}[A_k A_j] \]

Next, we separate the expectation of the product:

\[ \mathbb{E}[A_k]\mathbb{E}[1 - A_j] = \mathbb{E}[A_k] (\mathbb{E}[1] - \mathbb{E}[A_j]) \]

Since \(\mathbb{E}[1] = 1\), we have:

\[ \mathbb{E}[A_k] (1 - \mathbb{E}[A_j]) = \mathbb{E}[A_k] - \mathbb{E}[A_k] \mathbb{E}[A_j] \]

Putting it all together, we get:

\[ \text{Cov}(A_k, 1 - A_j) = (\mathbb{E}[A_k] - \mathbb{E}[A_k A_j]) - (\mathbb{E}[A_k] - \mathbb{E}[A_k] \mathbb{E}[A_j]) \]

Simplifying, we obtain:

\[ \text{Cov}(A_k, 1 - A_j) = \mathbb{E}[A_k] - \mathbb{E}[A_k A_j] - \mathbb{E}[A_k] + \mathbb{E}[A_k] \mathbb{E}[A_j] \]

\[ = - \mathbb{E}[A_k A_j] + \mathbb{E}[A_k] \mathbb{E}[A_j] \]

Since $\text{Cov}(A_k, A_j) = \mathbb{E}[A_k A_j] - \mathbb{E}[A_k] \mathbb{E}[A_j]$ :

\[ \text{Cov}(A_k, 1 - A_j) = -\text{Cov}(A_k, A_j) \]
\end{proof}

$L = -\left[ y \log(p) + (1 - y) \log(1 - p) \right]$

\section{Subsequence Clustering}\label{apdx:clustering}  
Recall, the objective of subsequence clustering is:
\begin{align}
\argmin{\mathbf{\Theta}, \tilde{\vy}} &\sum_{k=1}^{K} \Bigg[
 \sum_{y =1 }^{|Y|} \sum_{\vz_p \in \tilde{\vy}_k}  (\underbrace{-\ell\ell(\vz_p,\Theta_k)}_{\text{\normalfont log likelihood}} \\ \nonumber
 &+ \underbrace{\beta \mathds{1}\{\vz_{p-1} \not\in \tilde{\vy}_k\}}_{\text{\normalfont temporal consistency}} ) + \underbrace{\|  \lambda \odot \Theta_k\|_1}_{\text{\normalfont time-invariant representation}}\Bigg].
\end{align}

\subsection{Graphical lasso}\label{apdx:graphlasso}  
We use graphical lasso as a part of our model.
Given the mode-(\ndim+1) matricization of the time series \Norder{\ndim+1} \TTS,
$\matXN{\tensor}{\ndim+1} \in \mathbb{R}^{\timestep \times \probdim}$,
the graphical lasso
estimates the sparse Gaussian inverse covariance matrix (i.e., network) 
$\model \in \mathbb{R}^{\probdim \times \probdim}$,
also known as the precision matrix,
with which we can interpret
pairwise conditional independencies
among $\probdim$ variables,
e.g., if $\model_{i,j}=0$
then variables $i$ and $j$ are conditionally
independent given the values of all the other variables.
The optimization problem is given as follows:
\begin{align}\label{eq:gl}
\textrm{minimize}_{\model \in S^{p}_{++}}
    &\sparseparam||\model||_{od,1} - \sum_{t=1}^{\timestep} \loglike(\matXN{\tensor}{\ndim+1}_{t,},\model), \\
\loglike(\element,\model) = &-\frac{1}{2} (\element - \mu)^T \model (\element - \mu) \nonumber \\
    &+ \frac{1}{2} \log \textrm{det} \model -\frac{\probdim}{2} \log(2\pi) , 
\end{align}
where $\model$ must be a symmetric positive definite ($S^{p}_{++}$).
$\loglike(\element, \model)$ is the log-likelihood and $\mu \in \mathbf{R}^{\probdim}$ is the empirical mean of $\matXN{\tensor}{\ndim+1}$.
$\sparseparam \geq 0$ is a hyperparameter for determining
the sparsity level of the network,
and $\|\cdot\|_{od,1}$ indicates the off-diagonal \lonenorm.
Since \eq{\ref{eq:gl}} is a convex optimization problem,
its solution is guaranteed to converge to the global optimum with
the alternating direction method of multipliers (ADMM) \cite{admm}
and can speed up the solution time.
This is similar to expectation maximization (EM), with the two subroutines corresponding to the E and M steps, respectively.

\subsection{Cluster Assignment}
Given the model parameters, specifically the inverse covariances for each of the $K$ clusters, subsequence clustering involves assigning the $P$ subsequences, $X_1,\ldots,X_P$, to these $K$ clusters. The goal is to maximize the data likelihood while minimizing the number of cluster transitions across the EEG recordings.
This combinatorial optimization problem, which assigns $P$ points to $K$ clusters, has $K^{P}$ possible assignment combinations. Despite this complexity, we can achieve a globally optimal solution with only $O(KP)$ operations.
This is accomplished using a dynamic programming approach. This method is equivalent to finding the minimum cost Viterbi path for the length-$P$ sequence.

\subsection{Solving the Toeplitz Graphical Lasso}

After obtaining the clustering assignments, the M-step of our EM algorithm focuses on updating the inverse covariances based on the points assigned to each cluster. 
The next task is to update the cluster parameters $\{\Theta_1, \ldots, \Theta_K\}$ by opti while holding $\mathbf{P}$ constant. We can solve for each $\Theta_k$ in parallel.
To do so, we notice that we can rewrite the negative log likelihood in Problem Eq.(1) in terms of each $\Theta_k$. This likelihood can be expressed as 
\begin{align*}
\small
\sum_{X_p \in \tilde{\vy}_k} -\ell\ell(X_t,\Theta_k) = -|\tilde{\vy}_k|(\log\det\Theta_k + \mathrm{tr}(S_k \Theta_k)), 
\end{align*}
where $|\tilde{\vy}_k|$ is the number of points assigned to cluster $i$, $S_k$ is the empirical covariance of these points. Therefore, the M-step of our EM algorithm is 
\begin{align}
\small
&\mathrm{minimize}\qquad -\log\det\Theta_k + \mathrm{tr}(S_k \Theta_k) + \frac{1}{|\tilde{\vy}_k|}\|\lambda \circ \Theta_k\|_1 \nonumber\\
&\mathrm{subject\ to}\qquad \Theta_k \in \mathcal{T}.
\label{Toeplitz}
\end{align}
Problem Eq.(1) is a convex optimization problem, which we call the \emph{Toeplitz graphical lasso}. This is a variation on the well-known graphical lasso problem  where we add a block Toeplitz constraint on the inverse covariance.  
The original graphical lasso defines a tradeoff between two objectives, regulated by the parameter $\lambda$: minimizing the negative log likelihood, and making sure $\Theta_k$ is sparse.
When $S_k$ is invertible, the likelihood term encourages $\Theta_k$ to be near $S_k^{-1}$.
Our problem adds the additional constraint that $\Theta_k$ is block Toeplitz.
$\lambda$ is a $nw \times nw$ matrix, so it can be used to regularize each sub-block of $\Theta_k$ differently. 
Note that $\frac{1}{|\tilde{\vy}_k|}$ can be incorporated into the regularization by simply scaling $\lambda$; as such, we typically write Problem \eqref{Toeplitz} without this term (and scale $\lambda$ accordingly) for notational simplicity.

\end{document}